\documentclass[A4, 10 pt, conference]{ieeeconf}  

\IEEEoverridecommandlockouts                              

\usepackage{caption}
\usepackage[skins]{tcolorbox}  
\usepackage{subfigure}
\usepackage{multicol}

\usepackage{mathtools}
\usepackage{graphics} 
\usepackage{epsfig} 
\usepackage{mathptmx} 
\usepackage{times} 
\usepackage{amsmath} 
\usepackage{amssymb}  
\usepackage{graphicx}
\usepackage{epstopdf}
\usepackage{pdfpages}
\usepackage{cite}
\usepackage{amsbsy}
\usepackage{bm}
\usepackage{url}
\usepackage{algorithm}
\usepackage{algpseudocode}
\usepackage{accents}

\usepackage{fancyhdr}

\def\horizontaldistance{\kern2pt}

\title{\LARGE \bf
Kinematic Base State Estimation for Humanoid using Invariant Extended Kalman Filter}

\author{Amirhosein Vedadi$^{1}$, Aghil Yousefi-Koma$^{1}$, Masoud Shariat-Panahi$^{1}$,  Mahdi Nozari$^{1}$
\thanks{$^{1}$School of Mechanical Engineering, College of Engineering, University of Tehran,
Tehran, Iran.
        {\tt\small aykoma@ut.ac.ir}}%
        }%

\makeatletter
\newcommand*{\rom}[1]{\expandafter\@slowromancap\romannumeral #1@}
\makeatother
\begin{document}

\maketitle
\thispagestyle{empty}
\pagestyle{empty}

\begin{abstract}
This paper presents the design and implementation of a Right Invariant Extended Kalman Filter (RIEKF) for estimating the states of the kinematic base of the Surena V humanoid robot. The state representation of the robot is defined on the Lie group $SE_4(3)$, encompassing the position, velocity, and orientation of the base, as well as the position of the left and right feet. In addition, we incorporated IMU biases as concatenated states within the filter.

The prediction step of the RIEKF utilizes IMU equations, while the update step incorporates forward kinematics. To evaluate the performance of the RIEKF, we conducted experiments using the Choreonoid dynamic simulation framework and compared it against a Quaternion-based Extended Kalman Filter (QEKF).
The results of the analysis demonstrate that the RIEKF exhibits reduced drift in localization and achieves estimation convergence in a shorter time compared to the QEKF. These findings highlight the effectiveness of the proposed RIEKF for accurate state estimation of the kinematic base in humanoid robotics.
\end{abstract}
\thispagestyle{fancy}
\section{INTRODUCTION}
Humanoid and bipedal robots have become increasingly popular in recent years due to their ability to perform complex tasks in a variety of environments. Accurate estimation of the base position, velocity, and orientation of the robot is of paramount importance for ensuring successful operation, particularly when considering the inherent noise present in its sensor measurements.
The significance of accurate base position and velocity estimation is twofold. On one hand, knowing the precise position and velocity of the base allows us to compute the center of mass (CoM) position and velocity \cite{masuya2020review}, as well as the divergent component of motion, which plays a critical role in robot control \cite{jeong2019anklehipstep, abdolahnezhad2022online}. On the other hand, accurately determining the robot's position is essential for localization purposes and planning subsequent tasks.

The most straightforward approach for estimating the base of a humanoid robot involves utilizing its forward kinematics. In this method, it is assumed that the supporting foot experiences neither rotation nor slippage while in contact with the ground. By employing encoders on each joint and employing kinematic modeling of the robot, an estimation of the base can be obtained. However, the effectiveness of this method is hindered by various factors such as sensor noise, slip in the contact of the foot, and inaccuracies in the kinematic modeling, ultimately leading to less desirable outcomes.
\begin{figure}[ht]
	\centering 
	\includegraphics[width=0.442\textwidth]{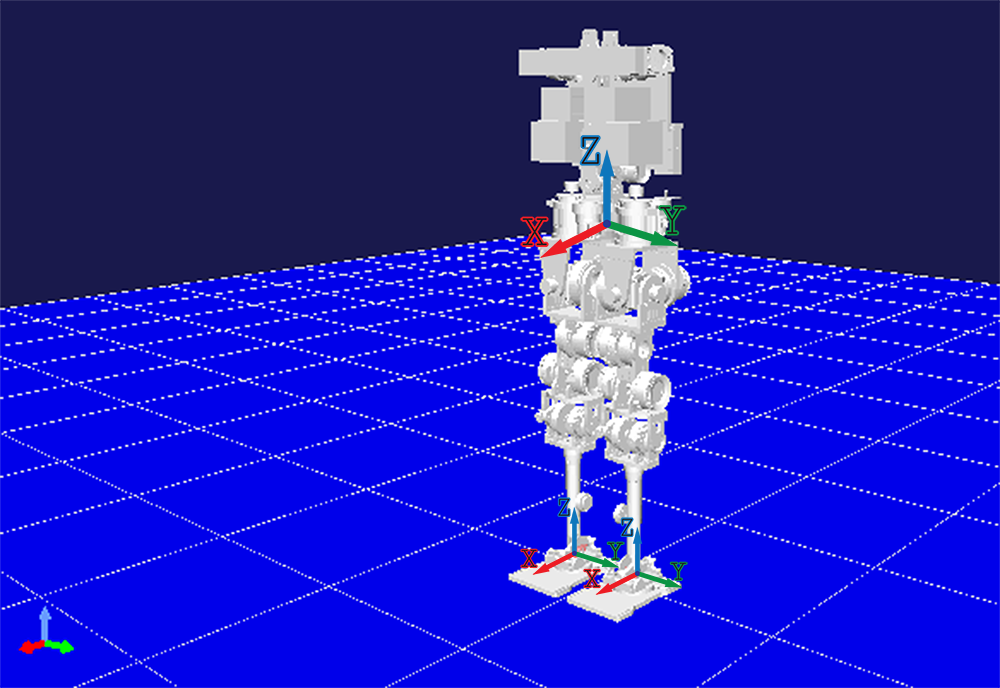}
	\caption{ Lower limb model of the Surena V robot in the Choreonoid simulator. Also, it illustrates the coordinate system for the kinematic base and feet.}
	\label{fig:surena_choreonoid}
\end{figure}
To address the aforementioned challenges, researchers have proposed combining data from various sensors available on humanoid robots, such as IMUs, with forward kinematics. Notably, Blouch et al. conducted significant research in this field \cite{bloesch2013state}. In their study, they employed IMU equations within the prediction step of an error state extended Kalman filter. A key contribution of their research was the inclusion of foot positions as state variables, facilitating the direct consideration of errors in leg slippage and movement. To model the movement of the foot in the contact equation, they utilized Brownian motion. In the update step of their filter, they incorporated kinematics equations specific to a quadruped robot. It was demonstrated that due to the unobservability of the yaw angle and absolute position of the base, drift in the filter results became unavoidable over time. Rotella et al. \cite{RotellaBRS14} proposed a similar filter to the aforementioned approach specifically designed for humanoid robots. Given that humanoid robots typically possess flat feet, any changes in their rotation can directly impact their position. To address this, the researchers included foot orientation as a state variable, allowing for its direct consideration in the estimation process. In another study by Falon et al. \cite{fallon2014drift}, a no-drift estimator was developed to address the issue of drift in humanoid robot localization. The researchers integrated IMU measurements, forward kinematics equations, and the output of a LIDAR point cloud to achieve accurate localization. To accomplish this, they employed a particle filter algorithm, which effectively localized the humanoid robot within a predefined map while mitigating the effects of drift.

In recent years, there has been an increasing demand to formulate estimation problems accurately in legged robots. This demand has led to the emergence of utilizing Lie groups theory and invariant estimator design as a significant development in this field. By leveraging the principles of Lie groups theory, researchers have been able to design estimators that are invariant, meaning they are robust and unaffected by transformations or changes in the robot's configuration. The utilization of Lie groups theory has resulted in improved uncertainty modeling for systems with states represented in Lie groups \cite{sola2017quaternion}. Bourmaod et al. \cite{bourmaud2015continuous} presented a framework for discrete and continuous state estimation utilizing an Extended Kalman Filter (EKF) with predict and update equations formulated in the context of Lie groups. In their paper, they introduced the concept of a concentrated Gaussian distribution, which defines a Gaussian distribution on both the Lie group and its equivalent Lie algebra. To facilitate computations, this distribution was transformed to Euclidean space using exponential mapping. The key advantage of this approach is that the resulting distribution in Euclidean space is no longer normal, thereby enabling improved performance in addressing nonlinear problems. 

Invariant filters are a class of filters where the estimation error of their state remains constant under different transformations applied to their state space \cite{bonnabel2005invariant}. Barau and Bonnabel \cite{barrau2016invariant} extended the concept of invariant filters to systems defined on Lie groups. These filters ensure that the state estimation error remains invariant under transformations in the Lie group space. In other words, these filters leverage the symmetry of the system to achieve a more robust formulation and enhance performance.

One notable consequence of using invariant filters is that if the system exhibits the group-affine property, the estimation error in the Lie algebra space follows linear differential equations, and the linearization process becomes independent of the estimated states. This characteristic of invariant filters contributes to improved convergence performance.

Hartley et al. \cite{Hartley2020} applied these invariant filters to effectively combine IMU equations with forward kinematics for accurate base estimation. Their work demonstrated that linear differential equations can describe the error dynamics in the Lie algebra space. Additionally, they highlighted the advantage that contrary to normal EKF approaches, the linearized system and update equations become independent of the estimated states. They also addressed the incorporation of IMU bias and foot positions as state variables in the filtering equations.

In a related study, Teng et al. \cite{teng2021legged} utilized a similar invariant filter to estimate the base in slippery environments. They dynamically adjusted the noise parameters online and incorporated forward kinematic velocity measurements in the update step to enhance the estimation accuracy.

Another application of these filters was demonstrated by Ramadoss et al. \cite{ramadoss2021diligent}, who considered the flat feet of a humanoid robot and applied the invariant filter to the iCub humanoid robot. They also conducted a comparative analysis of their filter with Rotella's filter, evaluating their performance in humanoid base estimation.

The primary objective of this paper is to introduce an approach for base estimation and localization of the Surena-V humanoid robot by designing an invariant filter. Notably, this study leverages the Choreonoid simulator \cite{nakaoka2012choreonoid}, to evaluate the performance of the proposed filter. The utilization of this simulator provides a realistic and robust platform for validating the effectiveness of the invariant filter. The results of this filter will be compared with a Quaternion-based Extended Kalman Filter (QEKF) \cite{bloesch2013state}.

In Section II, a concise introduction to Lie group theory will be provided, followed by a detailed description of the filter equations design. In Section III, the designed invariant filter will be tested using a Choreonoid simulation environment. Finally, in Section IV, the paper will be concluded by summarizing the findings and discussing the implications of the proposed invariant filter.
\section{Material AND Methods}
In this section, we provide a detailed account of the key concepts underlying the proposed methodology, including an overview of Lie groups and Lie algebras. Moreover, we delve into the explanation of the Invariant Extended Kalman Filter (InEKF), elucidating its distinct components and their respective roles within the designed filter.
\subsection{Overview of Lie Group and Lie Algebra}
Lie groups are mathematical groups characterized by their properties as smooth manifolds. Smooth manifolds, in turn, are topological spaces that possess differentiable properties at each point and exhibit local linearity.

If ($\pmb{\mathcal{X}}(t)$) represents a point on the Lie group ($\pmb{\mathcal{G}}$), the velocity ($\frac{\partial \pmb{\mathcal{X}}}{\partial t}$) of this point belongs to the tangent space of ($\pmb{\mathcal{G}}$) at this point ($\mathcal{T}_{\pmb{\mathcal{X}}} \mathcal{G}$). The structure of this tangent space remains consistent throughout the manifold due to its inherent smoothness. In particular, the tangent space at the identity element of the group ($\mathcal{T}_{\pmb{\mathcal{I}}} \mathcal{G}$) corresponds to the Lie algebra ($\mathfrak{g}$). It is worth noting that each Lie group is associated with a unique Lie algebra. 

In Lie algebra, the elements are represented as skew-symmetric matrices. These matrices can be transferred to the vector space of ($\mathbb{R}^m$) using the ($.^\vee$) operator, which transforms the matrix into a column vector ($m$ is Lie group's degree of freedom). Similarly, the ($.^\wedge$) operator converts elements from the vector space back to skew-symmetric matrices in the Lie algebra. To uniquely transform vectors from a Lie algebra ($\pmb{\xi}$) to a Lie group matrix ($\pmb{\mathcal{X}}$), the exponential transformation is utilized. This transformation maps elements of the Lie algebra to the corresponding elements of the Lie group. Conversely, the inverse of the exponential transformation is known as the logarithmic mapping.
\begin{gather}
\text{exp:}\quad \pmb{\mathcal{X}} = exp(\pmb{\xi}^\wedge)\\ 
\text{log:}\quad \pmb{\xi}^\wedge = log(\pmb{\mathcal{X}})
\end{gather}

Lie algebras can be locally defined at other points and subsequently transformed into a global Lie algebra using the adjoint transformation of the group. 
\begin{equation}\label{397}
Ad_{\pmb{\mathcal{X}}}(\pmb{\xi}^\wedge) = \pmb{\mathcal{X}}\pmb{\xi}^\wedge\pmb{\mathcal{X}}^{-1} \quad Ad_{\pmb{\mathcal{X}}}: \mathfrak{m} \to \mathfrak{m}
\end{equation}
Lie groups commonly used in robotics include ($SO(3)$) for three-dimensional rotations, ($SE(3)$) for combined rotations and translations, and ($SE_k(3)$) for extended rigid body transformations with deformations or kinematic changes. For further insights into this subject matter, please refer to the relevant resource \cite{sola2018micro}.

\subsection{Invariant Extended Kalman Filter}
In this section, we will provide a comprehensive description of our filter, its equation, and its respective components. Firstly, we will outline the state variables that form the foundation of our analysis. Secondly, we will delve into the prediction equations, which integrate the IMU measurements and foot contact state to forecast the kinematic base state of the humanoid robot. Lastly, we will elucidate the update equations, which leverage the forward kinematics of the robot to refine and improve the estimated base state based on sensor measurements.

\subsubsection{State Variables}
In the context of the kinematic base state estimation for humanoid robots, the state variables encompass several key components. These include the position ($\pmb{p}$), velocity ($\pmb{v}$), and rotation ($\pmb{R}$) of the kinematic base itself. Additionally, the positions of the left and right foot ($\pmb{p}^{c_L}$ and $\pmb{p}^{c_R}$) are also considered as distinct states within the estimation process. The inclusion of foot states is crucial as it enables the assessment of their respective errors and their impact on the accuracy of the base state estimation. The states are defined within the mathematical framework of $SE_4(3)$ as follows:
\begin{equation}
	\pmb{\mathcal{X}} = 
	\begin{bmatrix}
		\pmb{R} & \pmb{v} & \pmb{p} & \pmb{p}^{c_L} & \pmb{p}^{c_R}\\
		\pmb{0}_{1 \times 3}& 1 & 0 & 0 & 0 \\
		\pmb{0}_{1 \times 3}& 0 & 1 & 0 & 0 \\
		\pmb{0}_{1 \times 3}& 0 & 0 & 1 & 0 \\
		\pmb{0}_{1 \times 3}& 0 & 0 & 0 &  1
	\end{bmatrix}
	\in SE_4(3)
\end{equation}
The Lie algebra associated with the states can be expressed as follows:
\begin{equation}
	\pmb{\xi}^\wedge = 
	\begin{bmatrix}
		(\pmb{\xi}^R)^\wedge & \pmb{\xi}^v & \pmb{\xi}^p & \pmb{\xi}^{p_{c_L}} & \pmb{\xi}^{p_{c_R}}\\
		\pmb{0}_{1 \times 3}& 0 & 0 & 0 & 0 \\
		\pmb{0}_{1 \times 3}& 0 & 0 & 0 & 0 \\
		\pmb{0}_{1 \times 3}& 0 & 0 & 0 & 0 \\
		\pmb{0}_{1 \times 3}& 0 & 0 & 0 & 0
	\end{bmatrix}
\end{equation}

In our state representation, we do not incorporate IMU biases due to the absence of a Lie group that adheres to the group affine property of the system dynamics \cite{barrau2015non}. However, it is worth noting that an alternative approach can be employed by designing an incomplete invariant filter. This filter, while not fully satisfying the properties of an invariant filter, has been shown to yield improved results compared to other filters \cite{Hartley2020}. Consequently, we introduce the bias states in the following manner:
\begin{equation}
	\pmb{\theta} = vec(\pmb{b}^g, \pmb{b}^a)
\end{equation}
So the concatenated state representation of the system can be expressed as
$(\pmb{\mathcal{X}}, \pmb{\theta}) \in \mathcal{G} \times \mathbb{R}^6$,
where $\mathcal{G}$ represents the Lie group defined on $SE_4(3)$, and $\mathbb{R}^6$ represents the six-dimensional vector space associated with IMU biases states. 

\subsubsection{Prediction Equations}
To model the values of the IMU sensor, we employ the following equations:
\begin{gather} \label{eq:31} 
\tilde{\pmb{\omega}} = \pmb{\omega} + \pmb{w}^g + \pmb{b}^g\\
\tilde{\pmb{a}} = \pmb{a} + \pmb{w}^a + \pmb{b}^a
\end{gather}
Here, $\tilde{\pmb{\omega}}$ and $\tilde{\pmb{a}}$ represent the measured outputs of the IMU sensor, while $\pmb{\omega}$ and $\pmb{a}$ denote their respective actual values. The terms $\pmb{w}^g$ and $\pmb{w}^a$ represent the white noise components, while $\pmb{b}^g$ and $\pmb{b}^a$ represent the biases associated with the gyroscope and accelerometer, respectively. This particular modeling approach has been widely utilized in previous studies.
The prediction step of this filter can be formulated by the following equations:
\begin{gather} 
\label{eq:42}
\dot{\pmb{R}} = \pmb{R}(\tilde{\pmb{\omega}} - \pmb{b}^g - \pmb{w}^g)^\wedge = \pmb{R}(\bar{\pmb{\omega}} - \pmb{w}^g)^\wedge \\
\label{eq:43}
\dot{\pmb{v}} = \pmb{a} = \pmb{R}(\tilde{\pmb{a}}-\pmb{b}^a-\pmb{w}^a)+\pmb{g} = \pmb{R}(\bar{\pmb{a}}-\pmb{w}^a)+\pmb{g} \\
\label{eq:44}
\dot{\pmb{p}} = \pmb{v}\\
\dot{\pmb{p}}^{c_i} = \pmb{R} \pmb{R}_{fk}^{c_i} (-\pmb{w}^{c_i}), \quad i={L, R}\\
\dot{\pmb{b}}^g = \pmb{w}^{bg}\\
\label{eq:47}
\dot{\pmb{b}}^a = \pmb{w}^{ba}
\end{gather}
In equation (9), to derive the difference equation of rotation, we follow a similar approach as outlined in research \cite{sola2017quaternion}. The operator $()^\wedge$ denotes the skew-symmetric matrix operator. In equation (10), we initially perform a coordinate frame transformation from the IMU coordinate frame to the global coordinate frame. Subsequently, we eliminate the gravity vector ($\pmb{g}$) from the transformed equation. In equation (12), we model the dynamics of the contact foot using a Brownian model, incorporating a noise vector represented by $\pmb{w}^{c_i}$. The term $\pmb{R}_{fk}^{c_i}$ denotes the rotation matrix that describes the orientation of the foot relative to the base. Equations (13) and (14) describe the modeling of IMU biases as a Brownian motion process, incorporating noise vectors represented by $\pmb{w}^{bg}$ and $\pmb{w}^{ba}$, respectively. 

To implement these equations, it is necessary to employ the deterministic and discrete forms of the equations. For the discretization of the equations, we adopt the analytical integration technique, which has been employed in prior research studies such as \cite{Hartley2020, eckenhoff2017direct, sola2017quaternion}. 

One of the key distinctions of the Invariant Extended Kalman Filter (InEKF) lies in the method used to define the error for state variables. In this filter, the error for the first part of the states is defined as the right invariant error on a Lie group. The inverse of our states group can be expressed as follows:
\begin{equation}
	\pmb{\mathcal{X}}^{-1} = 
	\begin{bmatrix}
		\pmb{R}^T & -\pmb{R}^T\pmb{v} & -\pmb{R}^T\pmb{p} & -\pmb{R}^T\pmb{p}^{c_L} & -\pmb{R}^T\pmb{p}^{c_R}\\
		\pmb{0}_{4 \times 3} &  &  &\pmb{I}_{4 \times 4}
	\end{bmatrix}
\end{equation}
so the error is as follows:
\begin{equation}
\resizebox{0.485\textwidth}{!}{$
\begin{gathered}
\pmb{\eta} = exp(\pmb{\xi}^\wedge) = \bar{\pmb{\mathcal{X}}} \pmb{\mathcal{X}}^{-1} = \\
\begin{bmatrix}
\bar{\pmb{R}}\pmb{R}^T & \bar{\pmb{v}}-\bar{\pmb{R}}\pmb{R}^T\pmb{v} & \bar{\pmb{p}}-\bar{\pmb{R}}\pmb{R}^T\pmb{p} & \bar{\pmb{p}}^{c_L}-\bar{\pmb{R}}\pmb{R}^T\pmb{p}^{c_L} & \bar{\pmb{p}}^{c_R}-\bar{\pmb{R}}\pmb{R}^T\pmb{p}^{c_R}\\
\pmb{0}_{4 \times 3} &  &  &\pmb{I}_{4 \times 4}
\end{bmatrix}
\end{gathered}
$}
\end{equation}
With the inclusion of bias states, the concatenated error can be expressed as follows:
\begin{equation}
	\pmb{e} = (\pmb{\eta}, \pmb{\zeta}) = (\bar{\pmb{\mathcal{X}}} \pmb{\mathcal{X}}^{-1}, \bar{\pmb{\theta}} - \pmb{\theta})
\end{equation}
\begin{figure*}[h]
 \centering
 \subfigure[X-direction position]{\includegraphics[trim={2cm 9cm 2cm 9cm},clip, width=0.32\textwidth]{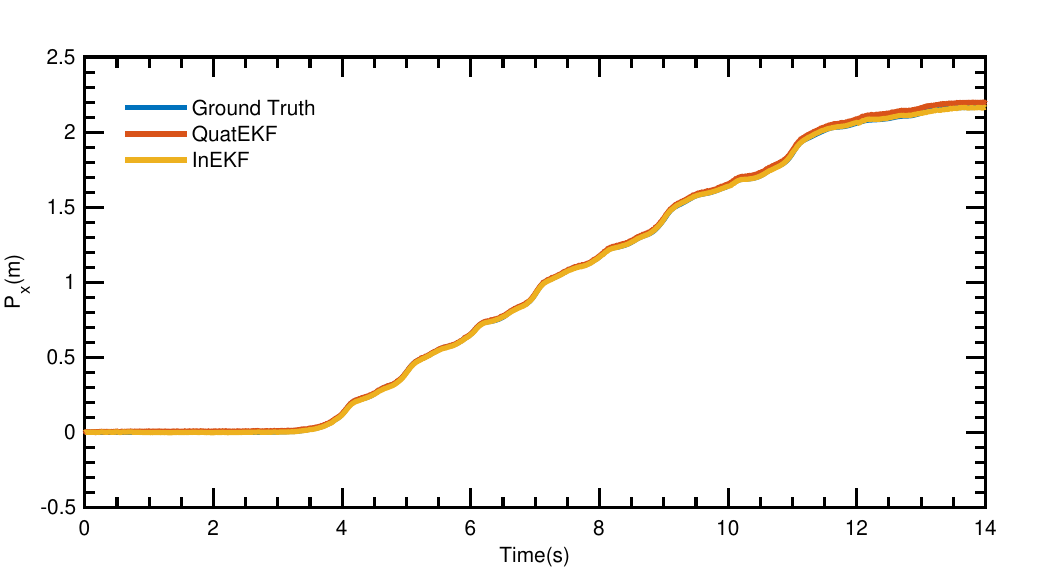}}
 \subfigure[Y-direction position]{\includegraphics[trim={2cm 9cm 2cm 9cm},clip, width=0.32\textwidth]{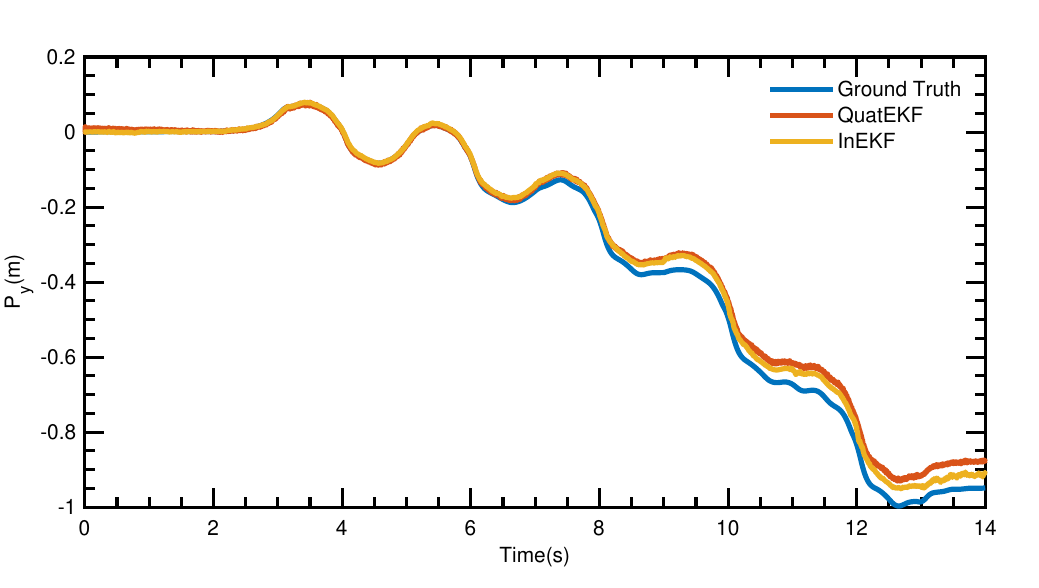}}
 \subfigure[Yaw angle]{\includegraphics[trim={2cm 9cm 2cm 9cm},clip, width=0.32\textwidth]{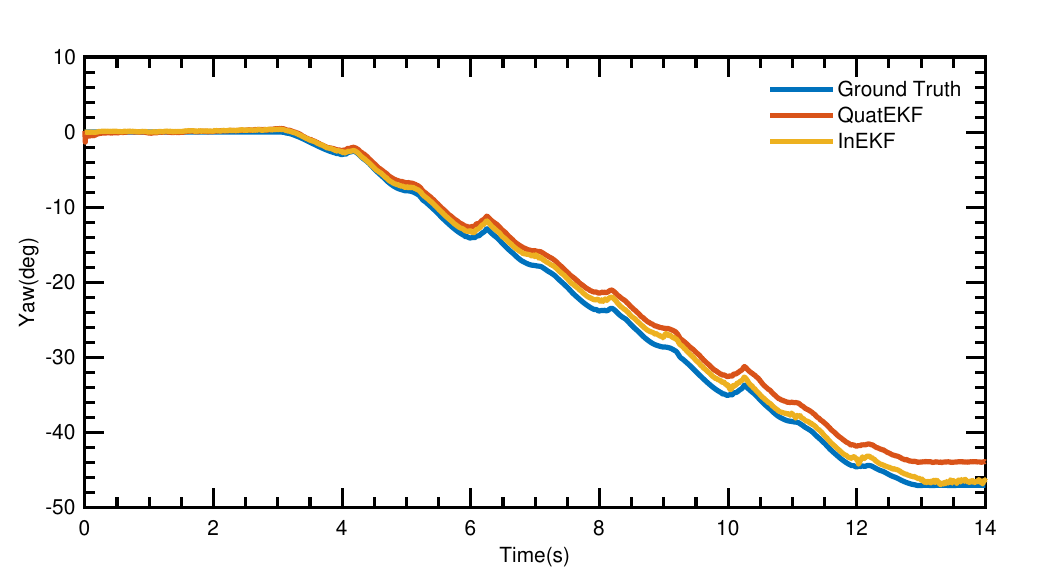}}
 \caption{Comparison of the output from both filters with the ground truth}
 \label{fig4}
\end{figure*}

\begin{figure*}[h]
 \centering
 \subfigure{\includegraphics[trim={2cm 9cm 2cm 9cm},clip, width=0.18\textwidth]{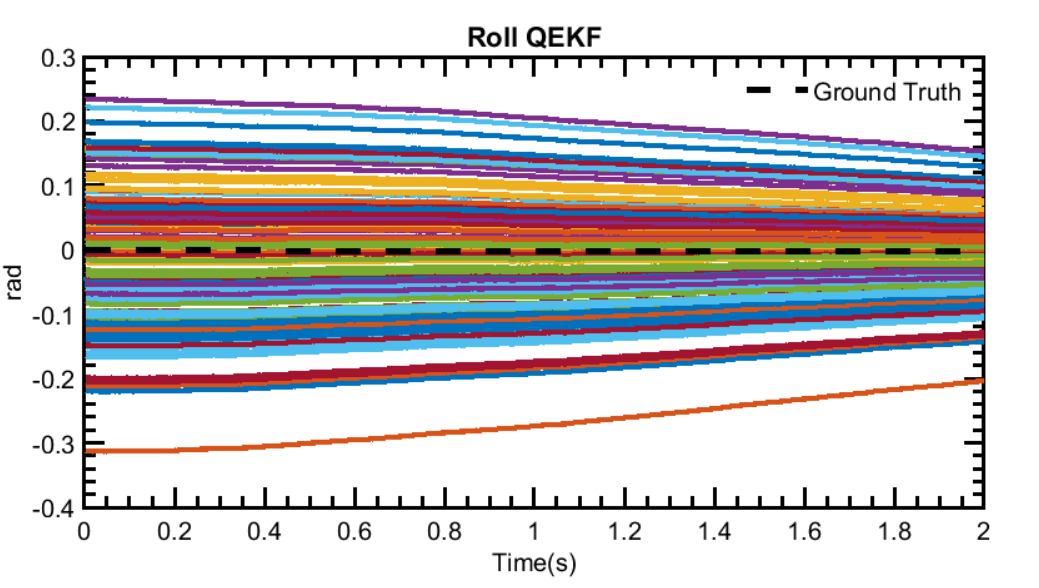}}
 \subfigure{\includegraphics[trim={2cm 9cm 2cm 9cm},clip, width=0.18\textwidth]{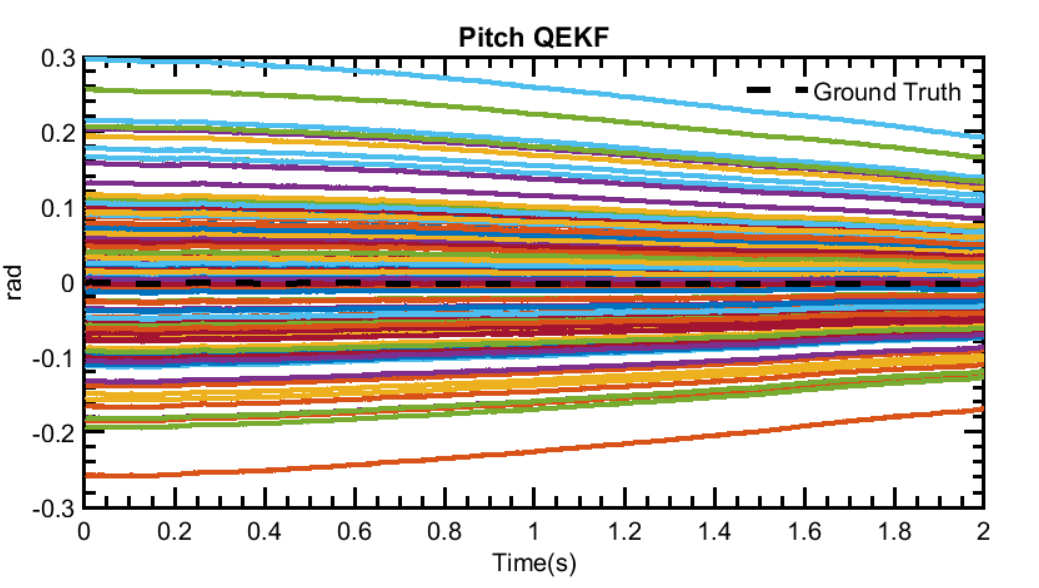}}
 \subfigure{\includegraphics[trim={2cm 9cm 2cm 9cm},clip, width=0.18\textwidth]{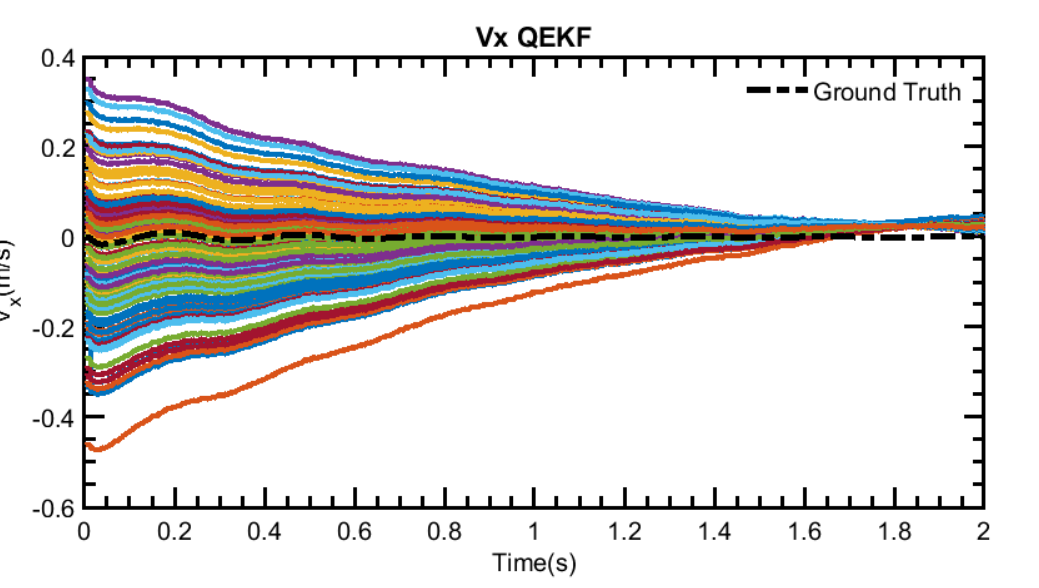}}
  \subfigure{\includegraphics[trim={2cm 9cm 2cm 9cm},clip, width=0.18\textwidth]{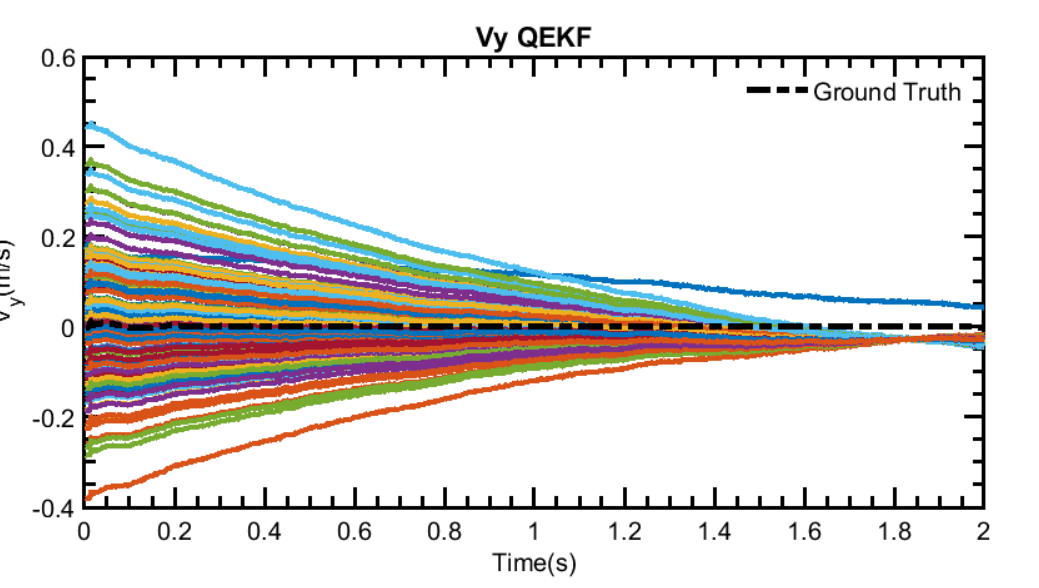}}
   \subfigure{\includegraphics[trim={2cm 9cm 2cm 9cm},clip, width=0.18\textwidth]{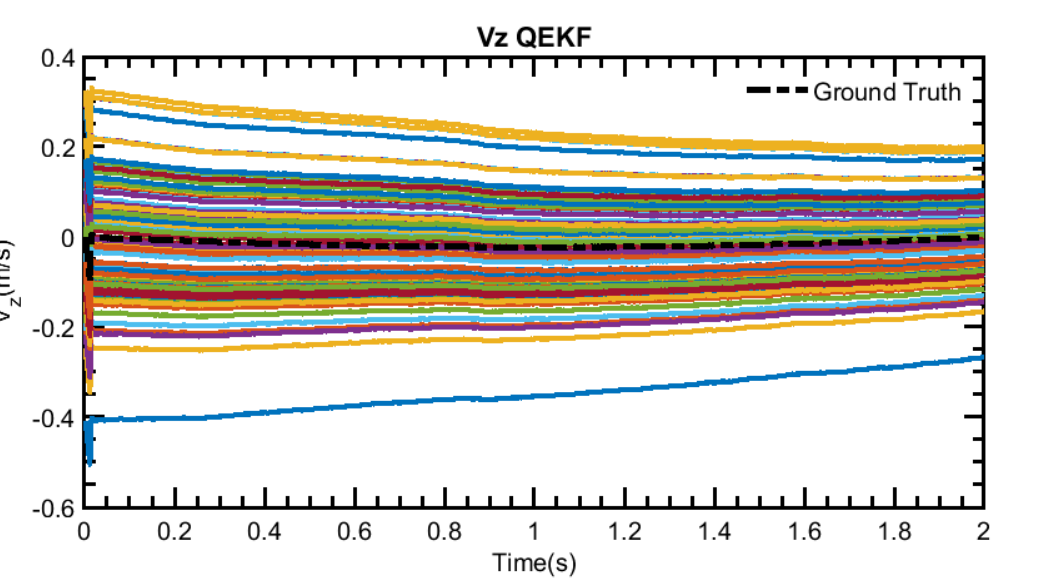}}
   \subfigure{\includegraphics[trim={2cm 9cm 2cm 9cm},clip, width=0.18\textwidth]{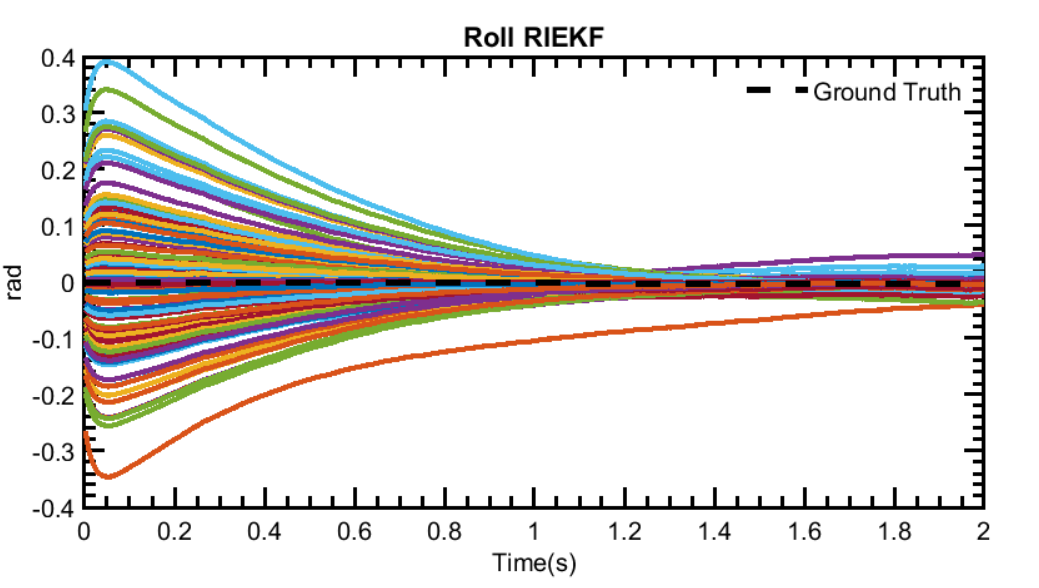}}
 \subfigure{\includegraphics[trim={2cm 9cm 2cm 9cm},clip, width=0.18\textwidth]{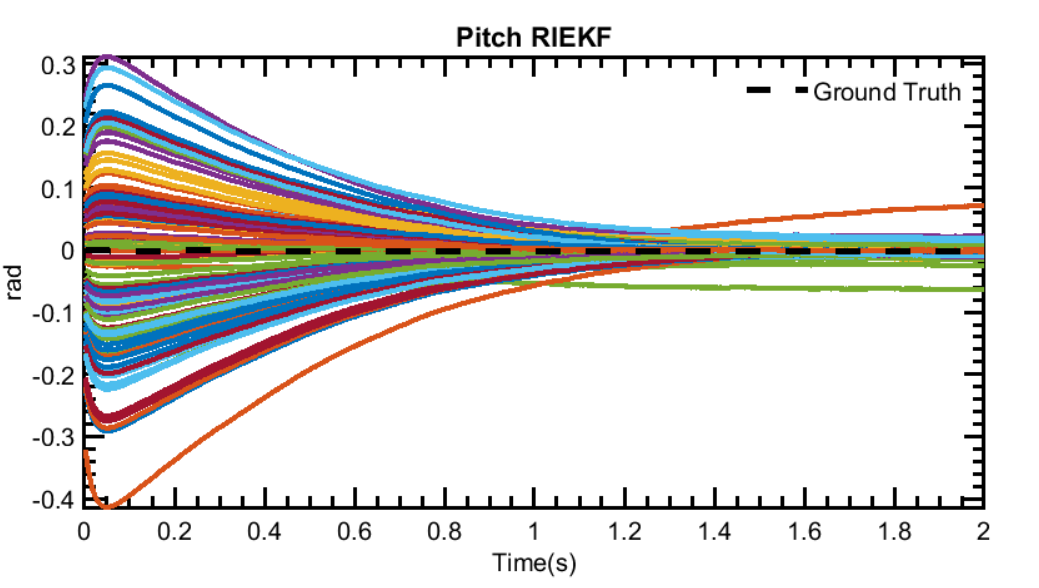}}
 \subfigure{\includegraphics[trim={2cm 9cm 2cm 9cm},clip, width=0.18\textwidth]{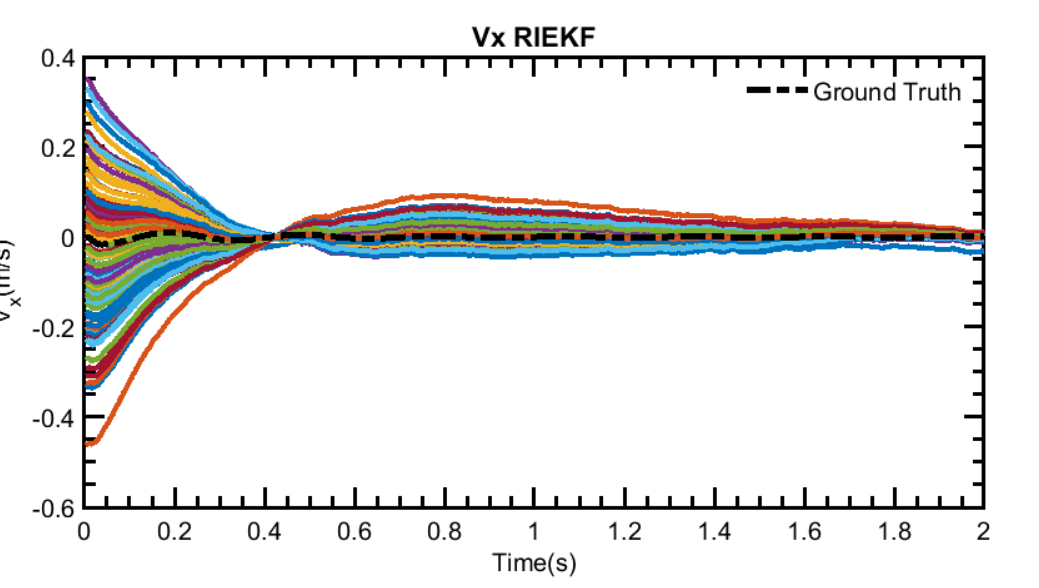}}
  \subfigure{\includegraphics[trim={2cm 9cm 2cm 9cm},clip, width=0.18\textwidth]{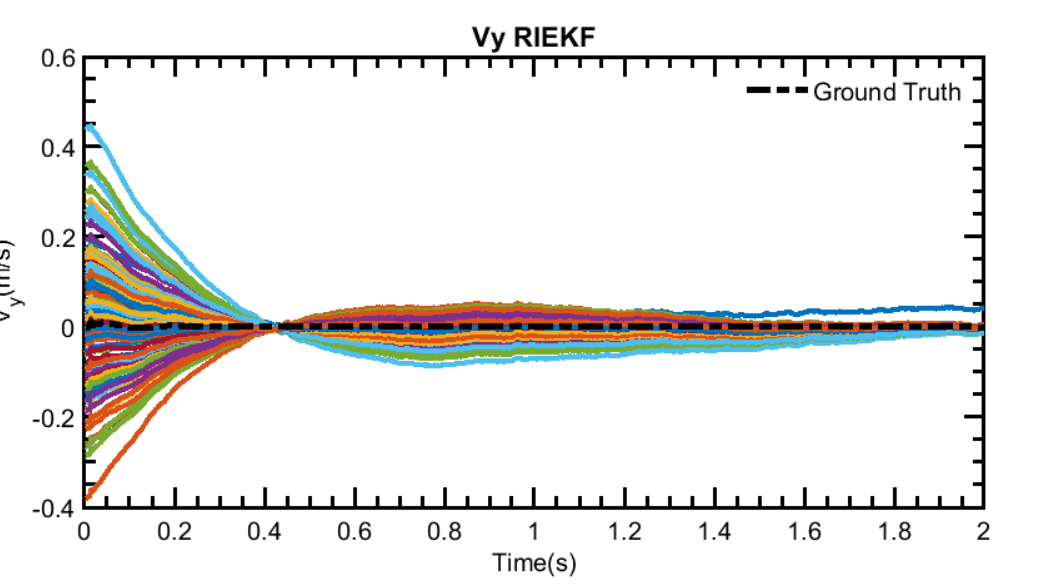}}
   \subfigure{\includegraphics[trim={2cm 9cm 2cm 9cm},clip, width=0.18\textwidth]{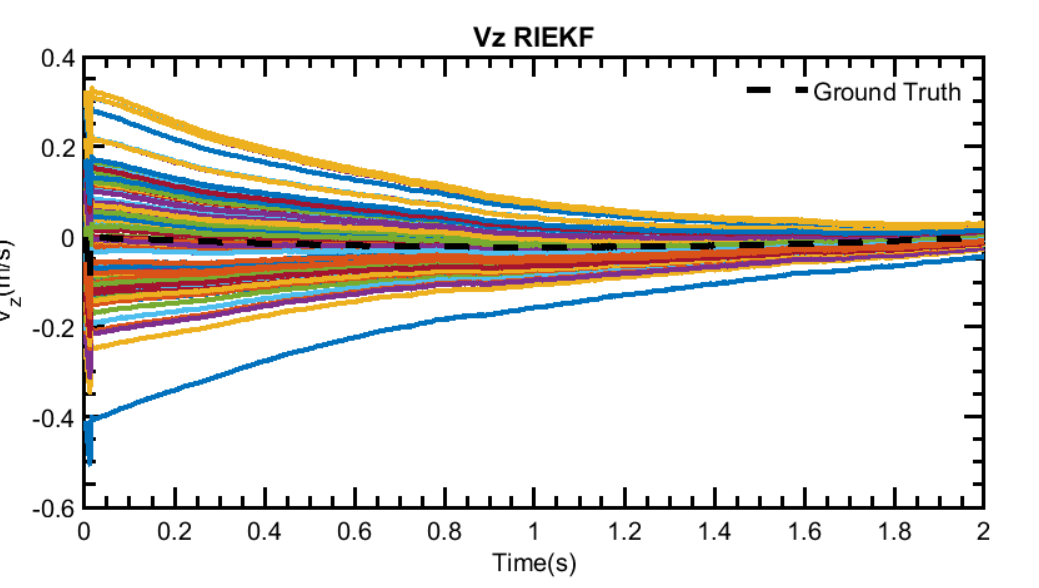}}
 \caption{Comparison of convergence for QEKF and RIEKF filters in the presence of noisy initial conditions of the states}
 \label{fig4}
\end{figure*}
To propagate the covariance in the prediction step, it is necessary to derive the error dynamics by differentiating equation (17). We refer to reference \cite{Hartley2020} to obtain the differentiation of the elements in equation (16). 
\begin{equation}
	\begin{matrix}
		\frac{d}{dt}(\bar{\pmb{R}}\pmb{R}^T) \approx (\pmb{R}(\pmb{w}^g - \pmb{\zeta}^g))^\wedge\\
		\frac{d}{dt}(\bar{\pmb{v}} - \bar{\pmb{R}}\pmb{R}^T \pmb{v}) \approx \pmb{g}^\wedge \pmb{\xi}^{R} + \pmb{v}^\wedge \pmb{R}(\pmb{w}^g - \pmb{\zeta}^g) + \pmb{R}(\pmb{w}^a - \pmb{\zeta}^a)\\
		\frac{d}{dt}(\bar{\pmb{p}} -\bar{\pmb{R}}\pmb{R}^T \pmb{p}) \approx \pmb{\xi}^{v} + \bar{\pmb{p}}^\wedge \pmb{R} (\pmb{w}^g - \pmb{\zeta}^g)\\
		\frac{d}{dt}(\bar{\pmb{p}^{c_i}} -\bar{\pmb{R}}\pmb{R}^T \pmb{p}^{c_i}) \approx
		(\bar{\pmb{p}^{c_i}})^\wedge \bar{\pmb{R}}(\pmb{w}^g - \pmb{\zeta}^g) + \pmb{R} \pmb{R}_{fk}^{c_i} \pmb{w}^{c_i}
	\end{matrix}
\end{equation}
By assuming that $\pmb{\eta} = exp(\pmb{\xi}^\wedge)$ can be approximated as $\pmb{I} + \pmb{\xi}^\wedge$, we can linearize the equation. Taking into account the bias terms, the resulting linearized equation will be:
\begin{equation}\label{eq:451}
	\frac{d}{dt}
	\begin{bmatrix}
		\pmb{\xi}\\
		\pmb{\zeta}
	\end{bmatrix} = 
	\pmb{F}
	\begin{bmatrix}
	\pmb{\xi}\\
	\pmb{\zeta}
	\end{bmatrix} + 
	\begin{bmatrix}
	Ad_{\bar{\pmb{\mathcal{X}}}} & \pmb{0}_{15 \times 6}\\
	\pmb{0}_{6 \times 15} & \pmb{I}_{6 \times 6}
	\end{bmatrix}
	\pmb{w}
\end{equation}
In this equation, $\pmb{F}$ and $\pmb{w}$ are:
\begin{equation*}
\resizebox{0.485\textwidth}{!}{$
\begin{matrix}
	\pmb{F} = 
	\begin{bmatrix}
		\pmb{0}_{3 \times 3} & \pmb{0}_{3 \times 3} & \pmb{0}_{3 \times 3} & \pmb{0}_{3 \times 3} & \pmb{0}_{3 \times 3} & -\bar{\pmb{R}} & \pmb{0}_{3 \times 3}\\
		\pmb{g}^\wedge & \pmb{0}_{3 \times 3} & \pmb{0}_{3 \times 3} & \pmb{0}_{3 \times 3} & \pmb{0}_{3 \times 3} & -\bar{\pmb{v}}^\wedge \bar{\pmb{R}} & -\bar{\pmb{R}}\\
		\pmb{0}_{3 \times 3} & \pmb{I}_{3 \times 1} & \pmb{0}_{3 \times 3} & \pmb{0}_{3 \times 3} & \pmb{0}_{3 \times 3} & -\bar{\pmb{p}}^\wedge \bar{\pmb{R}} & \pmb{0}_{3 \times 3}\\
		\pmb{0}_{3 \times 3}& \pmb{0}_{3 \times 3} & \pmb{0}_{3 \times 3} & \pmb{0}_{3 \times 3} & \pmb{0}_{3 \times 3} & -(\bar{\pmb{p}}^{c_L})^\wedge \bar{\pmb{R}} & \pmb{0}_{3 \times 3}\\
		\pmb{0}_{3 \times 3}& \pmb{0}_{3 \times 3}0 & \pmb{0}_{3 \times 3} & \pmb{0}_{3 \times 3} & \pmb{0}_{3 \times 3} & -(\bar{\pmb{p}}^{c_R})^\wedge \bar{\pmb{R}} & \pmb{0}_{3 \times 3}\\
		 &  &  & \pmb{0}_{6 \times 7} & & &
	\end{bmatrix}\\
 \\
	\pmb{w} = vec(\pmb{w}^g, \pmb{w}^a, \pmb{0}_{3 \times 1}, \pmb{R}_{fk}^{c_L}\pmb{w}^{c_L}, \pmb{R}_{fk}^{c_R}\pmb{w}^{c_R}, \pmb{w}^{bg}, \pmb{w}^{ba})
\end{matrix}$}
\end{equation*}
As evident in F, only the elements corresponding to noise and bias are dependent on the estimated variables. Hence, in a system without considering biases, the error would be independent of the estimated states. According to equation (19), the noise covariance matrix in continuous space can be represented as follows:
\begin{equation}
	\pmb{Q}_c = 
	\begin{bmatrix}
	Ad_{\bar{\pmb{\mathcal{X}}}} & \pmb{0}_{15 \times 6}\\
	\pmb{0}_{6 \times 15} & \pmb{I}_{6 \times 6}
	\end{bmatrix}
	Cov(\pmb{w})
	\begin{bmatrix}
	Ad_{\bar{\pmb{\mathcal{X}}}} & \pmb{0}_{15 \times 6}\\
	\pmb{0}_{6 \times 15} & \pmb{I}_{6 \times 6}
	\end{bmatrix}^T
\end{equation}
Regarding $Cov(\pmb{w})$, we have the following expression:
\begin{equation}\label{eq:426}
	Cov(\pmb{w}) = blkdiag(\pmb{Q}^g, \pmb{Q}^a, \pmb{0}_{3 \times 3}, \pmb{Q}^{c_L}, \pmb{Q}^{c_R}, \pmb{Q}^{bg}, \pmb{Q}^{ba})
\end{equation}
The operator blkdiag(.) represents the formation of a block diagonal matrix. The matrices $\pmb{Q}^g, \pmb{Q}^a, \pmb{Q}^{bg}$, and $\pmb{Q}^{ba}$ can be understood as block diagonal matrices with elements representing the square of the standard deviations of the noise in angular velocity, linear acceleration, angular velocity bias, and linear acceleration bias, respectively. Regarding $\pmb{Q}^{c_i}$, it is important to note that when one of the robot's legs is not in contact with the ground, its prediction equation is temporarily eliminated. During this state, the noise covariance associated with that leg is assigned a high value to increase uncertainty. Once the leg returns to contact with the ground, this value is decreased.
\begin{equation}\label{eq:427}
	\pmb{Q}^{c_i} = \pmb{R}_{fk}^{c_i} (\pmb{Q}^c_i + (1 - contact(c_i)) \times 10^5)(\pmb{R}_{fk}^{c_i})^T, i = L, R
\end{equation}
The covariance matrix $\pmb{Q}^c_i$ represents the noise covariance associated with the contact foot. The function $contact(.)$ is used to determine whether the foot is in contact with the ground or not. 

Once $\pmb{F}$ and $\pmb{Q}_c$ are determined, they need to be discretized. To achieve this, we employ the following approximations:
\begin{gather}
	{\pmb{F}}_k = exp(\pmb{F} \Delta t) = \pmb{I} + \pmb{F} \Delta t\\
\pmb{Q}_k = {\pmb{F}}_k \pmb{Q}_c {\pmb{F}}_k^T \Delta t
\end{gather}
After obtaining the discretized matrices, we can proceed to propagate the covariance of states using the following equation:
\begin{equation}\label{eq:430}
{\pmb{P}}_{k+1|k} = \pmb{F}_k \pmb{P}_{k|k} \pmb{F}_k^T + \pmb{Q}_k
\end{equation}
\subsubsection{Update Equations}
In the update step, we will utilize the forward kinematics equations of the robot. The primary equation can be expressed as follows:
\begin{equation}\label{eq:431}
_B{}\pmb{p}^{c_i} = \pmb{R}^T(\pmb{p}^{c_i} - \pmb{p}) + \pmb{w}^{c_i}
\end{equation}
To leverage the outcomes of the InEKF, we express this equation in the form of right invariant observations ($\pmb{Y} = \pmb{\mathcal{X}}^{-1} \pmb{b} + \pmb{V}$), as suggested in reference \cite{barrau2015non}:
\begin{equation}
    \resizebox{0.485\textwidth}{!}{$
    \underbrace{
        \begin{bmatrix}
            _B{}\pmb{p}^{c_i}\\ 
            0\\ 
            1\\ 
            -1\\
            0
        \end{bmatrix}
    }_{\pmb{Y}} = 
    \underbrace{
        \begin{bmatrix}
            \pmb{R}^T & -\pmb{R}^T\pmb{v} & -\pmb{R}^T\pmb{p} & -\pmb{R}^T\pmb{p}^{c_L} & -\pmb{R}^T\pmb{p}^{c_R}\\
            \pmb{0}_{4 \times 3} &  &  &\pmb{I}_{4 \times 4}
        \end{bmatrix}
    }_{{\pmb{\mathcal{X}}^{-1}}}
    \underbrace{
        \begin{bmatrix}
        \pmb{0}_{3 \times 1}\\ 
        0\\ 
        1\\ 
        -1\\
        0
        \end{bmatrix}
    }_{{\pmb{b}}} + 
    \underbrace{
        \begin{bmatrix}
        \pmb{w}^{c_i}\\ 
        0\\ 
        0\\ 
        0\\
        0
        \end{bmatrix}
    }_{{\pmb{V}}}
    $}
\end{equation}
therefore, the update equations for states can be formulated as follows:
\begin{equation}\label{eq:358}
\hat{\pmb{\mathcal{X}}} = exp(\pmb{L}(\bar{\pmb{\mathcal{X}}} \pmb{Y} - \pmb{b})) \bar{\pmb{\mathcal{X}}}
\end{equation}
In this equation, $\pmb{L}$ represents the gain matrix, $\hat{\pmb{\mathcal{X}}}$ is updated states and $\bar{\pmb{\mathcal{X}}} \pmb{Y} - \pmb{b}$ denotes the innovation term. The update equation for the right invariant error can be expressed as follows:
\begin{equation}
    \hat{\pmb{\eta}} = \hat{\pmb{\mathcal{X}}} \pmb{\mathcal{X}}^{-1} = exp(\pmb{L}(\pmb{\eta} \pmb{b} - \pmb{b} + \bar{\pmb{\mathcal{X}}} \pmb{V}))\pmb{\eta}
\end{equation}
Upon careful examination of the equation (28), it can be observed that the first four rows of $\pmb{b}$ are zero. By computing the subsequent four rows of $\bar{\pmb{\mathcal{X}}} \pmb{Y} - \pmb{b}$, it can be deduced that they will also be zero. Hence, it is possible to utilize a reduced-order gain and a selector matrix, denoted as $\pmb{K}$, and
$\pmb{\Pi} = blkdiag(\pmb{I}_{3 \times 3}, \pmb{0}_{3 \times 4})$, and rewrite equation as follows:
\begin{equation}
	\begin{matrix}
		\hat{\pmb{\mathcal{X}}} = exp(\pmb{K} \pmb{\Pi} \bar{\pmb{\mathcal{X}}} \pmb{Y}) \bar{\pmb{\mathcal{X}}}
	\end{matrix}
\end{equation}

To update the covariance matrix of states, we need to linearize equation (29) using the approximation
$\pmb{\eta} = exp(\pmb{\xi}^\wedge) = \pmb{I} + \pmb{\xi}^\wedge$. This linearization process yields the following result, as described in reference \cite{Hartley2020}:
\begin{equation}
	\hat{\pmb{\xi}} = \pmb{\xi} - \pmb{K}(
	\pmb{H}^{c_i} \pmb{\xi} - \bar{\pmb{R}} \pmb{w}^{c_i})
\end{equation}
In this equation, the H matrix for the left and right leg can be represented as follows:
\begin{equation}
	\begin{matrix}
		\pmb{H}^{c_L} =
		\begin{bmatrix}
			\pmb{0}_{3 \times 3}&\pmb{0}_{3 \times 3}& -\pmb{I}_{3 \times 3} &\pmb{I}_{3 \times 3} &\pmb{0}_{3 \times 3}
		\end{bmatrix}\\
		\pmb{H}^{c_R} =
		\begin{bmatrix}
		\pmb{0}_{3 \times 3}&\pmb{0}_{3 \times 3}& -\pmb{I}_{3 \times 3} & \pmb{0}_{3 \times 3}&\pmb{I}_{3 \times 3}
		\end{bmatrix}
	\end{matrix}
\end{equation}
Now, following the standard Kalman filter approach, we can update the covariance matrix using the following equations:
\begin{equation}
\begin{matrix}
	\pmb{N} = \pmb{R}Cov(\pmb{w}^{c_i}) \pmb{R}^T\\
	\pmb{S} = \pmb{H}^{c_i} \pmb{P} (\pmb{H}^{c_i})^T + \pmb{N}\\
	\pmb{K} = \pmb{P} (\pmb{H}^{c_i})^T \pmb{S}^{-1}\\
	\pmb{P}_{k+1|k+1} = (\pmb{I} - \pmb{K} \pmb{H}^{c_i}) \pmb{P} (\pmb{I} - \pmb{K} \pmb{H}^{c_i})^T + \pmb{K} \pmb{N} \pmb{K}^T
\end{matrix}
\end{equation}
These equations are independent of IMU biases. When considering IMU biases, we include two 3x3 zero matrices at the end of the matrix H. Additionally, when both of the robot's feet are in contact with the ground, we employ the following concatenated matrices for them:
\begin{equation}
    \resizebox{0.485\textwidth}{!}{$
	\begin{matrix}
		\pmb{Y} =
		\begin{bmatrix}
			\pmb{Y}^L\\
			\pmb{Y}^R
		\end{bmatrix},
		\pmb{b} = 
		\begin{bmatrix}
		\pmb{b}^L\\
		\pmb{b}^R
		\end{bmatrix},
		\begin{bmatrix}
		\pmb{H}^L\\
		\pmb{H}^R
		\end{bmatrix},\\
		\pmb{N} = blkdiag(\pmb{N}^L, \pmb{N}^R), 
		\pmb{\Pi} = blkdiag(\pmb{\Pi}, \pmb{\Pi}), 
		\bar{\pmb{\mathcal{X}}} = blkdiag(\bar{\pmb{\mathcal{X}}}, \bar{\pmb{\mathcal{X}}})
	\end{matrix}
    $}
\end{equation}

\section{Simulations and Results}
In this section, a comparative analysis will be conducted between the results obtained from the right invariant filter and the quaternion-based filter \cite{bloesch2013state}.

\subsection{System Overview}
For the simulation of the designed filters, we employed the Choreonoid dynamic simulator (Figure 1), coupled with ROS to extract data from the simulation for the filters. This simulator has been widely utilized for dynamic simulations of humanoid robots \cite{jeong2019robust, caron2019stair}. In our study, we utilized the lower limb model of the SURENA V robot, which is equipped with an IMU sensor on its base. Additionally, the robot's joints are equipped with encoders, allowing us to determine the forward kinematics of the robot based on its geometric properties. Contact detection was performed by leveraging the trajectory planning data of the robot.

The robot executed turning movements with a step length of 17 cm and an angle of 0.15 radians. Each step time was 1 second, with a double support phase of 0.1 seconds (equivalent to a speed of 0.61 km/h). The trajectory planning for the walking motion in this study is based on the research presented in reference \cite{vedadi2021bipedal}. The robot's control system operated at a frequency of 500 Hz. To enhance realism, Gaussian noise was introduced to the output of the IMU and forward kinematics. The standard deviations of these noise sources are presented in Table I.
\begin{table}[]
	\centering
	\label{table:41}
	\caption{Standard Deviations of Noise Added to the Simulation Data}
	\begin{tabular}{cc}
		\hline
		Parameter                         & Value \\ \hline
		angular velocity & $0.05 \frac{rad}{s}$  \\
        linear acceleration & $0.015 \frac{m}{s^2}$  \\
	    forward kinematics & $0.002 m$  \\ \hline
	\end{tabular}
\end{table}

\begin{table}[]
	\centering
	\label{table:42}
	\caption{Standard Deviations of Sensor Noise and Initial States Noise for Both Filters Incorporated in the Filtering Equations}
	\begin{tabular}{cccc}
		\hline
		    Sensor Parameter& Value  &Initial State Parameter&Value  \\ \hline
		         angular velocity & $0.05 \frac{rad}{s}$ &orientation & $0.1 rad$   \\
	             linear acceleration & $0.08 \frac{m}{s^2}$  &velocity  & $0.15 \frac{m}{s}$  \\
		       angular velocity bias  & $0.001 \frac{rad}{s^2}$ & position             & $0.1 m$   \\
		        linear acceleration bias    & $0.001 \frac{m}{s^3}$ &    feet position       & $0.1 m$   \\
		        forward kinematic & $0.05 m$  & angular velocity bias & $0.2 \frac{rad}{s}$ \\
		      contact foot velocity & $0.1 \frac{m}{s}$   &               linear acceleration bias& $0.2 \frac{m}{s^2}$   \\ \hline
	\end{tabular}
\end{table}
\begin{figure*}[htbp]
 \centering
 \subfigure{\includegraphics[trim={2cm 9cm 2cm 9cm},clip, width=0.48\textwidth]{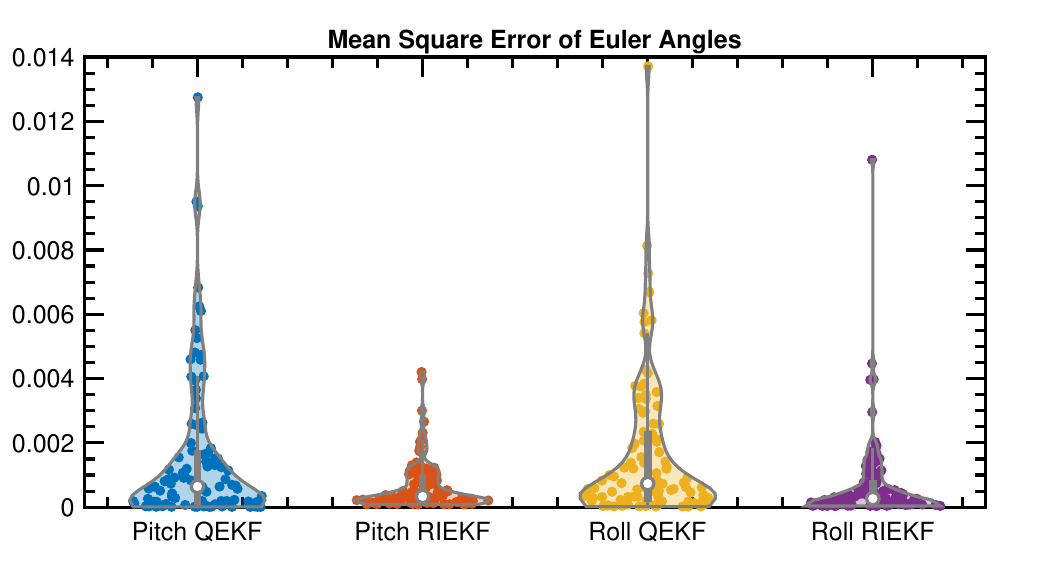}}
 \subfigure{\includegraphics[trim={2cm 9cm 2cm 9cm},clip, width=0.48\textwidth]{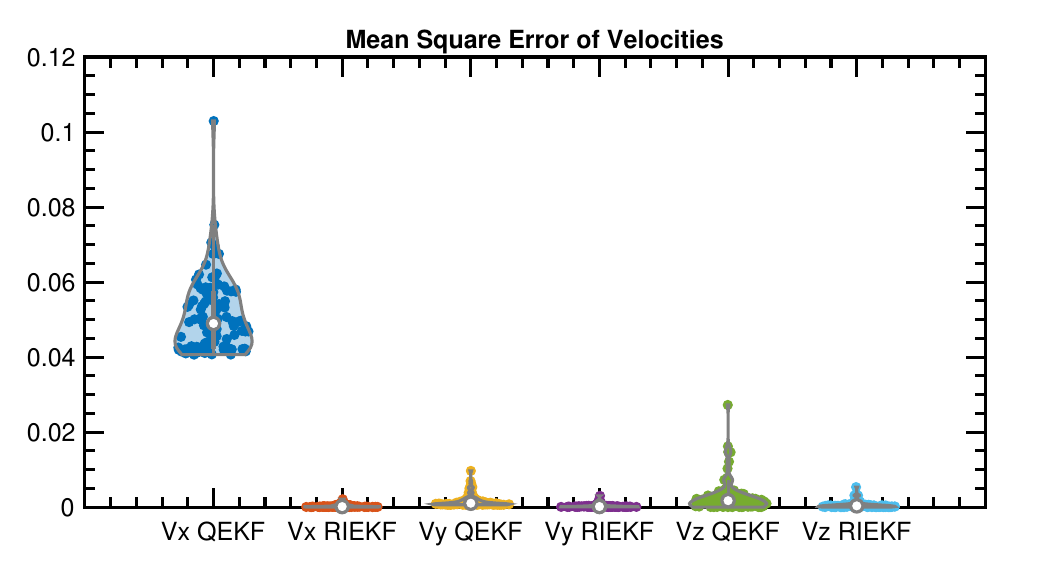}}
 \caption{Distribution of mean squared error (MSE) for different observable states}
 \label{fig4}
\end{figure*}
\subsection{Evaluation of Localization}
During the robot's walking, both filters were employed to estimate the kinematic base states. The standard deviations listed in Table II were used to represent the uncertainty in the initial state and sensor noises in the formulas of both filters. As highlighted in previous studies \cite{Hartley2020, bloesch2013state}, accurately estimating the base position and yaw angle poses challenges in terms of observability. In this study, we specifically examine the estimation of these states during walking. Figure 2 presents the results of the estimated values obtained from our approach, compared against the ground truth values derived from simulation computations. The comparison reveals that the RIEKF exhibits less drift than the QEKF along the robot's path, as evident from the results. Additionally, the mean squared error of the estimated states was computed by comparing them with the ground truth data, as presented in Table III. 

\begin{table}[]
	\centering
	\label{table:41}
	\caption{Mean Square Error of $p_x$, $p_y$ and $yaw$ for QEKF and RIEKF methods}
	\begin{tabular}{cccc}
		\hline
		Filter& $p_x (m)$ & $p_y(m)$ & $yaw(rad)$ \\ \hline
		RIEKF & 0.0000097 &0.0006052 &0.0002286  \\
	    QEKF & 0.0002555 &0.0014 & 0.0012 \\ \hline
	\end{tabular}
 
\end{table}

\subsection{Evaluation of Convergence}
In order to evaluate the convergence time of the two filters for observable states, a series of experiments was conducted. The states were randomly initialized 100 times from a normal distribution, with a standard deviation corresponding to the values reported in Table II. Subsequently, the filters were executed using these initial conditions, and the resulting outputs were visualized in Figure 3. The figures clearly demonstrate that the RIEKF filter consistently achieves convergence earlier than the QEKF filter.

Additionally, the mean squared error was calculated between the filter results and the ground truth obtained from the simulation for the 100 conducted tests, and their distribution was analyzed. The distribution of these MSE values is illustrated in Figure 4 using a violin plot. It is noteworthy that the RIEKF filter consistently exhibits lower MSE values compared to the QEKF filter for all states.

\section{Conclusion}
In this research, we developed a Right Invariant Extended Kalman Filter (RIEKF) for performing base state estimation in a humanoid robot. The states of the filter were defined on a Lie group $SE_4(3)$. We used IMU equations for the prediction step and forward kinematics for the update step. 
To validate the performance of the RIEKF, we conducted a comparative analysis with a Quaternion-based Extended Kalman Filter (QEKF) using the Choreonoid dynamic simulation platform. The results demonstrated that the RIEKF outperformed the QEKF in terms of reduced drift for unobservable states and faster convergence time for observable states.
The findings of this research highlight the effectiveness of the RIEKF in improving base state estimation for humanoid robots.

Looking ahead, future work will focus on incorporating visual filters to complement the RIEKF and further improve state estimation accuracy. By combining the RIEKF with visual-based filters, we aim to reduce drift over longer periods of time, enhancing the overall robustness and reliability of the state estimation system in humanoid robotics applications.

\addtolength{\textheight}{-8.7cm}   


\bibliography{bibliography}
\bibliographystyle{IEEEtran}

\end{document}